\documentclass[letterpaper, 10 pt, conference]{ieeeconf}  % Comment this line out if you need a4paper

\IEEEoverridecommandlockouts                              % This command is only needed if 
\usepackage[T1]{fontenc}
\usepackage[latin1]{inputenc}
\usepackage{amsfonts}
\usepackage{mathtools}
\usepackage{booktabs}
\usepackage{subcaption}
\usepackage{graphicx} % for pdf, bitmapped graphics files

\title{
\small \vspace{-3em}
Cite as: L. Philipp, F. M. L{\'o}pez, and J. Triesch, ``Embodiment Shapes Rolling Behavior in a Multimodal Infant Model'', in \textit{2026 IEEE International Conference on Development and Learning (ICDL)}. IEEE, 2026, pp. 1-7. \\[1em]
\LARGE \bf
Embodiment Shapes Rolling Behavior in a Multimodal Infant Model
}

\author{Leon Philipp$^{1}$, Francisco M. L\'{o}pez$^{2}$, Jochen Triesch$^{3}$% <-this % stops a space
    \thanks{This work was supported by the Deutsche Forschungsgemeinschaft (German Research Foundation, DFG) under Germany's Excellence Strategy (EXC 3066/1 ``The Adaptive Mind'', Project No. 533717223). J.T. was supported by the Johanna Quandt foundation. We thank Miles Lenz for his work on an earlier version of the rolling simulations and Matej Hoffmann for granting permission to include the frames of the infant from Fig.~\ref{fig:roll_illustrations}.}%
	\thanks{$^{1}$Leon Philipp is with the Frankfurt Institute for Advanced Studies, Germany, and with the Goethe University Frankfurt, Germany. {\tt\footnotesize philipp@fias.uni-frankfurt.de}}%
    \thanks{$^{2}$Francisco M. L\'{o}pez is with the Frankfurt Institute for Advanced Studies, Germany, and with the University of New South Wales, Australia. {\tt\footnotesize lopez@fias.uni-frankfurt.de}}
    \thanks{$^{3}$Jochen Triesch is with the Frankfurt Institute for Advanced Studies, Germany, and with the Goethe University Frankfurt, Germany. {\tt\footnotesize triesch@fias.uni-frankfurt.de}}
}

\begin{document}

\maketitle
\thispagestyle{empty}
\pagestyle{empty}

%%%%%%%%%%%%%%%%%%%%%%%%%%%%%%%%%%%%%%%%%%%%%%%%%%%%%%%%%%%%%%%%%%%%%%%%%%%%%%%%
\begin{abstract}
Rolling over is one of the earliest milestones in infant motor development, reflecting the emergence of coordinated, whole-body sensorimotor control. Here, we conduct a computational study of infant rolling using MIMo, a virtual infant embodiment equipped with proprioception and vestibular sensation. MIMo learns supine-to-prone rolls with reinforcement learning. Interestingly, the learned behaviors capture developmental trends and coordination patterns consistent with those reported in real infants, including improved performance and faster execution with age. Our results explain how infant capabilities and constraints can give rise to realistic behaviors in artificial agents, with a particular emphasis on how motor development is shaped by the changing body morphology. This work highlights the role of embodied computational models as a powerful tool for studying sensorimotor development.
\end{abstract}

%%%%%%%%%%%%%%%%%%%%%%%%%%%%%%%%%%%%%%%%%%%%%%%%%%%%%%%%%%%%%%%%%%%%%%%%%%%%%%%%
\section{INTRODUCTION}

Early motor development is characterized by the emergence of sensorimotor behaviors in a rapidly growing body \cite{adolph2017development,franchak2024update}. Rolling is one of the first major motor milestones, requiring the coordinated control of the whole body. Infants learn to roll between 2 and 7 months of age \cite{adolphmotordevelopment}, once they can lift their heads in the prone position and before sitting and crawling. Rolling has been described extensively in observational studies \cite{kobayashi16,mcgraw,posturalexperience}. Common analysis techniques include the classification of movement patterns in infants \cite{kobayashi16} and in adults \cite{richter89}, ground pressure analysis \cite{kobayashi2021}, and studies on the effect of the mechanical environment, such as rolling on an incline \cite{siegel25incline1,siegel24incline2}. However, the mechanisms underlying the emergence of rolling remain difficult to isolate due to experimental limitations with infant studies. By way of example, Siegel et al. \cite{siegel24:muscleactivation} performed a challenging study of the muscle activations involved in roll overs using electromyography (EMG), but only measured the muscles from the legs and torso, not the arms and head.

Some of these experimental limitations can be overcome by making use of computational models as a complementary methodology. A promising approach is to rely on models and artificial agents that can explain the interplay between different components of development beyond the scope of experimental studies. In particular, embodied agents with realistic physics simulations offer an interesting framework for learning and analyzing behaviors. Rolling has been comparatively understudied in computational developmental science despite its status as one of the earliest whole-body milestones. To the best of our knowledge, there are no existing computational models recreating infant rolling in artificial agents, although roll overs have been considered as a first milestone in a locomotion curriculum for simulated robots \cite{meng2023rolling}. On the other hand, humanoid robots have been used to explore rolling patterns on pressure sensitive mats \cite{sowande2024design} as well as motion kinematics \cite{kanagawa2018rolling}. However, little attention has been given to the specific role of the embodiment in the learning of this behavior.

\begin{figure}[!b]
    \centering
    \begin{subfigure}{\columnwidth}
    \centering\includegraphics[width=0.24\columnwidth]{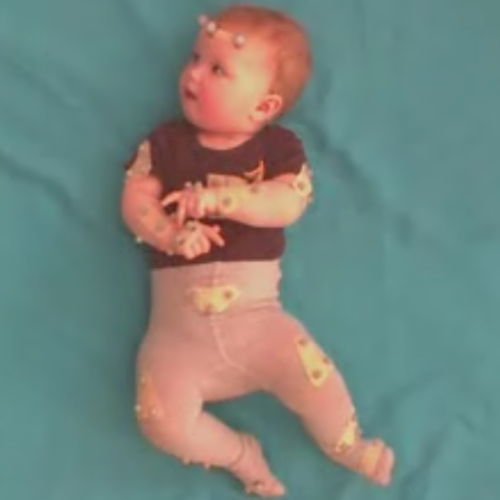}
    \centering\includegraphics[width=0.24\columnwidth]{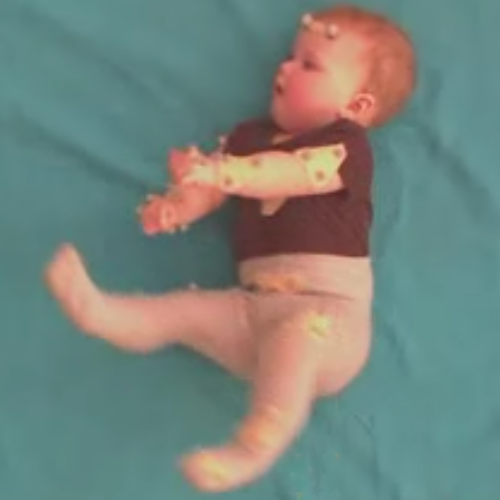}
    \centering\includegraphics[width=0.24\columnwidth]{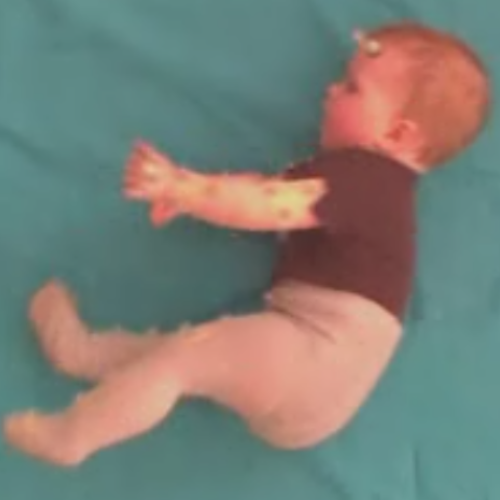}
    \centering\includegraphics[width=0.24\columnwidth]{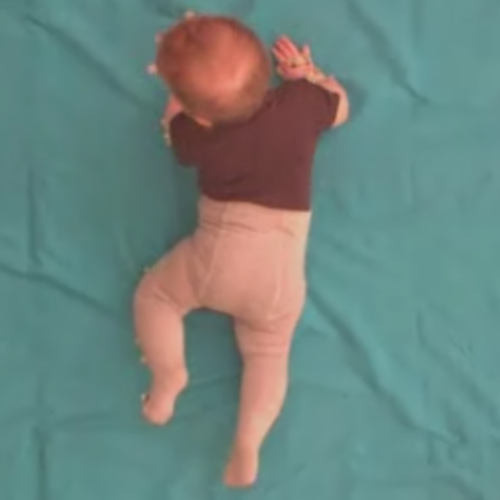}
    \caption{8-month-old infant.}
    \end{subfigure}
    
    \vspace*{1em}
    
    \centering
    \begin{subfigure}{\columnwidth}
    \centering\includegraphics[width=0.24\columnwidth]{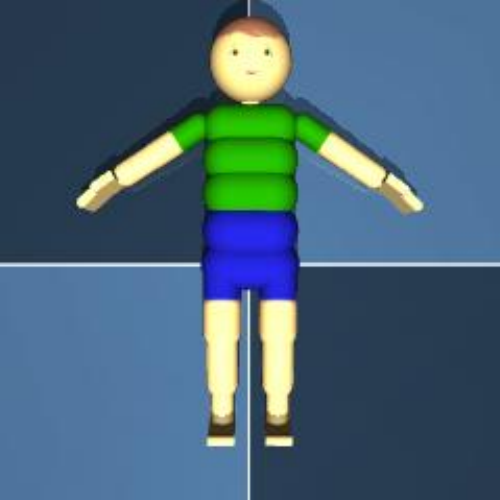}
    \centering\includegraphics[width=0.24\columnwidth]{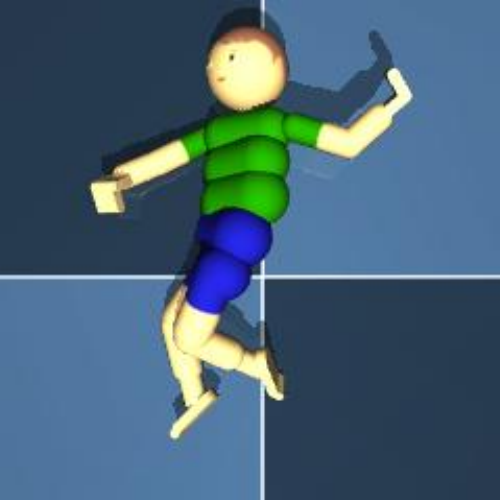}
    \centering\includegraphics[width=0.24\columnwidth]{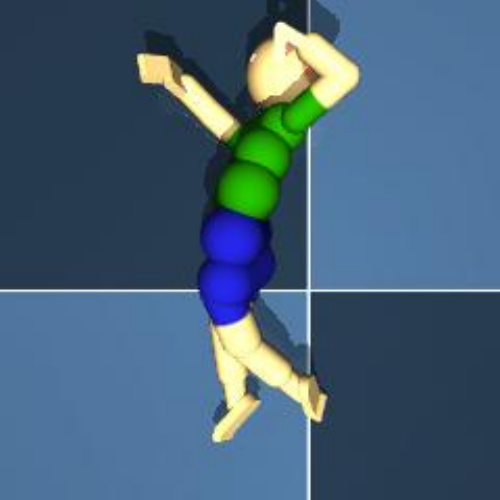}
    \centering\includegraphics[width=0.24\columnwidth]{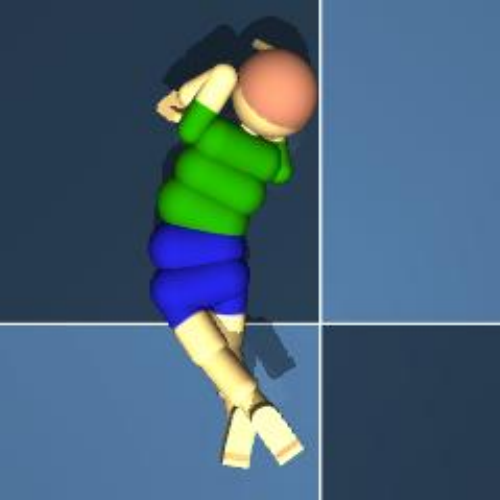}
    \caption{9-month-old MIMo \cite{mimo}.}
    \end{subfigure}
    \caption{Supine-to-prone rolling in real and simulated infants.}
    \label{fig:roll_illustrations}
\end{figure}

In this work, we present a computational model of infant rolling using MIMo \cite{mimo}, a multimodal embodied infant model (see Fig.~\ref{fig:roll_illustrations}). Our model combines a complex physical body with proprioceptive and vestibular sensing. Rolling behavior is actively learned through reinforcement learning, allowing MIMo to autonomously discover how to roll given his bodily constraints. We investigate two key aspects of rolling. First, we explore the developmental embodiment progression by varying MIMo's body between the ages of 1 and 9 months while keeping all other training components unchanged. Second, we analyze the resulting behaviors using metrics proposed in experimental psychology, including coordination patterns and muscle activations. Our work highlights the potential of embodied computational models as complementary tools for studying sensorimotor development.

\begin{figure*}[!t]
    \begin{subfigure}{0.24\textwidth}
        \centering\includegraphics[width=\textwidth]{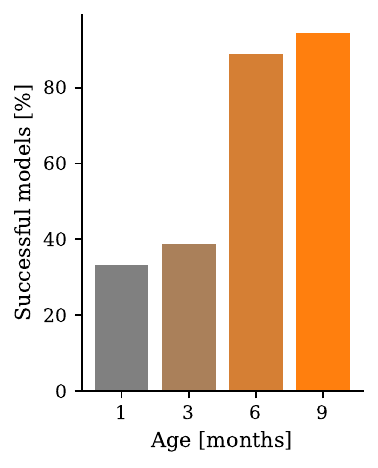}
        \caption{Success rates.}
    \end{subfigure}
    \begin{subfigure}{0.24\textwidth}
        \centering\includegraphics[width=\textwidth]{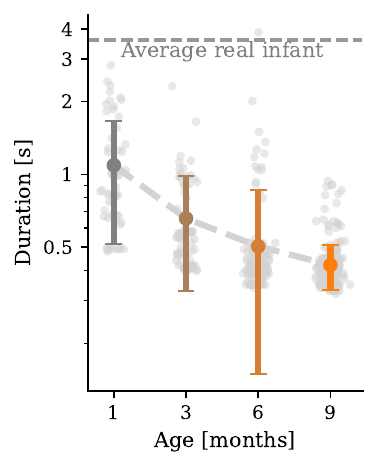}
        \caption{Durations.}
    \end{subfigure}
    \begin{subfigure}{0.48\textwidth}
        \centering\includegraphics[width=\textwidth]{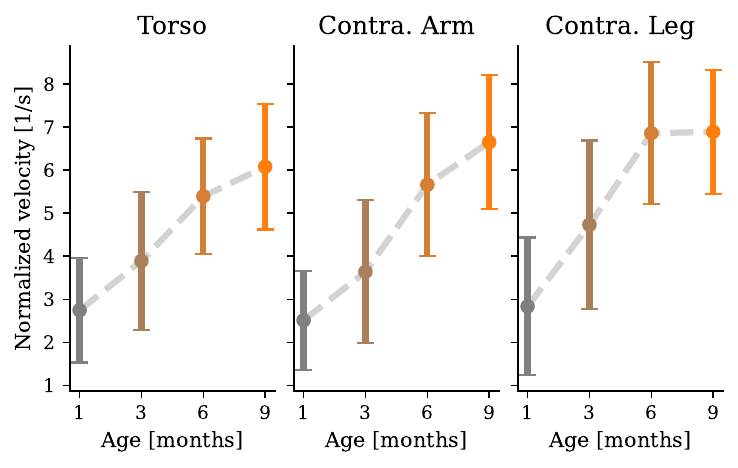}
        \caption{Normalized velocities.}
    \end{subfigure}
    \caption{Supine-to-prone rolling statistics for embodiments of different ages. (a) Success rates are measured over 40 test episodes. A model is deemed successful if it achieves a roll over in at least \(75\%\) of those episodes, i.e., 30 out of 40. (b) Roll durations are measured from the start of an episode until reaching a side-lying (lateral) position. The roll duration of real infants is the average from 24 infants \cite{siegel24:muscleactivation}. (c) The normalized velocity is calculated by fitting limb displacements to a normalized sigmoid curve and taking the velocity at the inflection point. 36 models were trained for the age of 9 months and 18 models each for ages 1, 3, and 6 months. Error bars represent one standard deviation.}
    \label{fig:supine2prone}
\end{figure*}

\section{METHODS}

\subsection{Environment}

We use MIMo \cite{mimo}, a virtual infant model designed to capture fundamental aspects of infant sensorimotor development. His age can be varied between birth and 2 years \cite{mimogrows}. MIMo is physically grounded in the MuJoCo simulation platform, which allows us to simulate the consequences of body contacts, gravity, and movement. MIMo is equipped with multimodal sensing, in particular proprioception and a vestibular system. Proprioception provides information about the current joint angles, velocities, torques, and control commands, while vestibular sensation measures the orientation and motion of the head. MIMo can actuate each of his joints, amounting to a total of 46 degrees-of-freedom (DOF), with a spring-damper model.

The environment consists of MIMo lying on a flat surface in a random starting position. Supine-to-prone rolls are analyzed most throughout the literature \cite{kobayashi16, mcgraw, siegel24:muscleactivation, richter89} and thus are our main focus here. The emergence of rolling happens between 2 and 7 months \cite{adolphmotordevelopment} and experimental studies typically study infants between the ages of 6 and 9 months \cite{kobayashi16,kobayashi2021,siegel24:muscleactivation}. To match this and also explore the developmental trajectories, we use 4 different embodiment ages: 1, 3, 6, and 9 months. 

\begin{figure*}[!t]
    \centering\includegraphics[width=\textwidth]{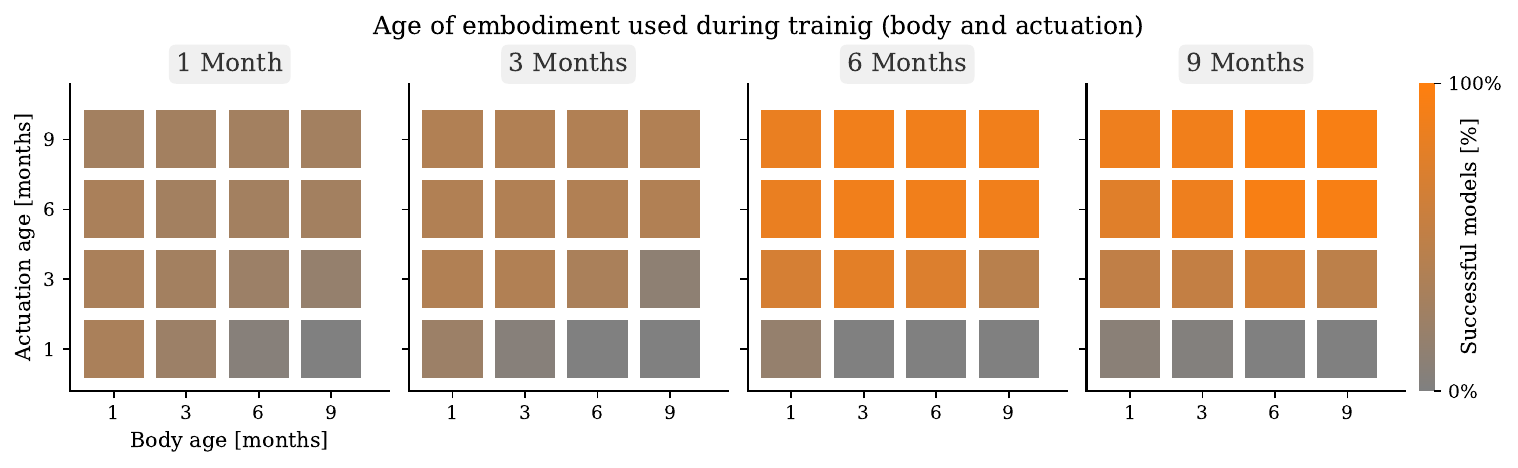}
    \caption{Disentangled cross-embodiment evaluation. For each pair of body and actuation age, models are tested for 40 episodes with different initial conditions. See Fig.~\ref{fig:supine2prone} for details of model success and number of models trained.}
    \label{fig:dcee}
\end{figure*}

\subsection{Training}

MIMo learns to roll over through reinforcement learning. He must learn a policy that maps the multimodal sensory inputs to continuous commands for all joints. To do so, he is trained to maximize an extrinsic densely-shaped reward function \cite{liu2022goal} and a large sparse reward upon rolling success (see Section \ref{sec:reward}) with Proximal Policy Optimization (PPO) \cite{ppo}. We use the default implementation from the Stable-Baselines3 library \cite{stable-baselines3}. 

Each training episode lasts a maximum of 500 steps, each step consisting of a physics simulation of \(10~\mathrm{ms} \), with early stopping if the roll over is successful. The total training time consists of \(10^6\) steps (corresponding to under 3 hours of practice). All the training procedures are repeated with multiple random seeds. 

\subsection{Roll progression}

To evaluate the transition from supine to prone, we use a metric based on the orientation of the torso and pelvis relative to the direction of gravity. In the supine position, both the torso and pelvis face upwards, opposite to the direction of gravity. In the prone position, the opposite is true. We measure the achieved shift of orientation from the starting position to the goal position. In a supine-to-prone roll, the goal is always aligned with the direction of gravity. To quantify this, we compute the dot product of the local normalized sagittal axes (the axes pointing forward) between pelvis and torso and the global gravitational axis. This dot product gives us a continuous value $\rho_i\in[-1,1]$ ($i\in\{\mathrm{pelvis},\mathrm{torso}\}$).

%To evaluate the transition from supine to prone position, we use a continuous metric based on the orientation of MIMo's body. Specifically, we monitor the alignment of the local sagittal axes (the axes that is facing forwards) of the pelvis and torso relative to the global coordinate system.

%We denote these local sagittal axes as $v_{\text{pelvis}}$ and $v_{\text{torso}}$ for pelvis and torso, respectively. For the supine-to-prone roll, in the default supine position, these axes point vertically upward (aligned to the global positive z-axis $+z_{\text{glob}}=(0,0,+1)^T$). In the prone position, they are aligned with the global negative z-axis $-z_{\text{glob}}=(0,0,-1)^T$.

%To quantify the roll progression, we compute the dot product between MIMo's local vectors and the target vector, i.e. the negative global z axis for a supine-to-prone roll:
%\begin{equation}
%\rho_i = <v_i, -z_{\text{glob}}>,
%\end{equation}
%with $i\in\{\text{pelvis}, \text{torso}\}$. This projection gives us a value $\rho\in[-1,1]$ with
%\begin{itemize}
%\item $\rho=-1$: MIMo is in the supine position.
%\item $\rho=1$: MIMo is in a prone position.
%\end{itemize}

Finally, we average both to obtain a single orientation metric:
\begin{equation}
\Bar{\rho} = \frac{\rho_{\text{pelvis}} + \rho_{\text{torso}}}{2}.
\end{equation}

Additionally, we use $\Bar{\rho} \geq 0.95$ as a boolean condition for a successful roll. A half-roll, in which a baby lies on its side, is identified with the condition $\Bar{\rho} \geq 0.5$.

\subsection{Reward function} \label{sec:reward}

A successful roll yields a large sparse reward. However, sparse rewards are notoriously challenging in large action spaces, such as the roll over environment used here. To address this, we use a densely-shaped reward function \cite{liu2022goal} generated with Potential Based Reward Shaping (PBRS) \cite{pbrs}. We use the potential function $\Phi(s) = \Bar{\rho}$, and scale the potential difference by a factor of 100. Finally, we incorporate metabolic costs that penalize MIMo for excessive usage of his actuators. This penalization is computed as the squared sum of the control values $c$ and scaled by a factor $\alpha\geq0$. Put together, the reward function is expressed as:
\begin{equation} \label{eq:reward}
    r_t = \left\{
        \begin{array}{ l l }
        500 & \textrm{if } \Bar{\rho} \geq 0.95 \\
        100 \cdot (\overline{\rho_{t+1}} - \overline{\rho_t}) - \alpha \sum c^2 & \textrm{otherwise}
    \end{array}
    \right.
\end{equation}

The penalization factor \(\alpha\) is set to \(0.02\) by default. Different values are compared in the Appendix.

\subsection{Evaluations}

After training, the deterministic policy is frozen and used for evaluations. Unless specified otherwise, we average results over 40 test episodes with different initial body configurations. We also average across multiple runs of the same training condition with different random seeds. %The different evaluation procedures are described next.

\subsubsection{Success rate}

A model is deemed successful if it completes a roll over in at least \(75\%\) of the \(40\) test episodes. The success rate is computed as the fraction of successful models per condition. Besides the success rate, we also keep track of the roll duration, defined as the time from the start of an episode until reaching a side-lying (lateral) position, chosen to ensure direct comparability with the measured durations of real infant rolling reported by Siegel et al. \cite{siegel24:muscleactivation}.

\subsubsection{Body velocities}

By using a simulator like MIMo, we can easily keep track of the movement statistics of the entire body. Here, we choose to recreate the analysis reported by Kobayashi et al. \cite{kobayashi16}. To do this, we attach MuJoCo ``sites'' at the wrists, ankles and torso. Contrary to the experimental counterpart, the markers are never obscured and thus there is no need for secondary markers. The velocities of each limb are calculated by differentiating the measured displacement of each marker. Since each limb must cover a different distance during a roll over, Kobayashi et al.\ also use a normalized velocity. They propose to fit the total displacement to a sigmoid function and normalize it to amplitude 1. The normalized velocity is then taken as the slope at the center point, in units of \(1/\mathrm{time}\). The body parts tracked are the torso (TR), ipsilateral arm (IA), contralateral arm (CA), ipsilateral leg (IL), and contralateral leg (CL).

\begin{figure*}
    \includegraphics[width=\textwidth]{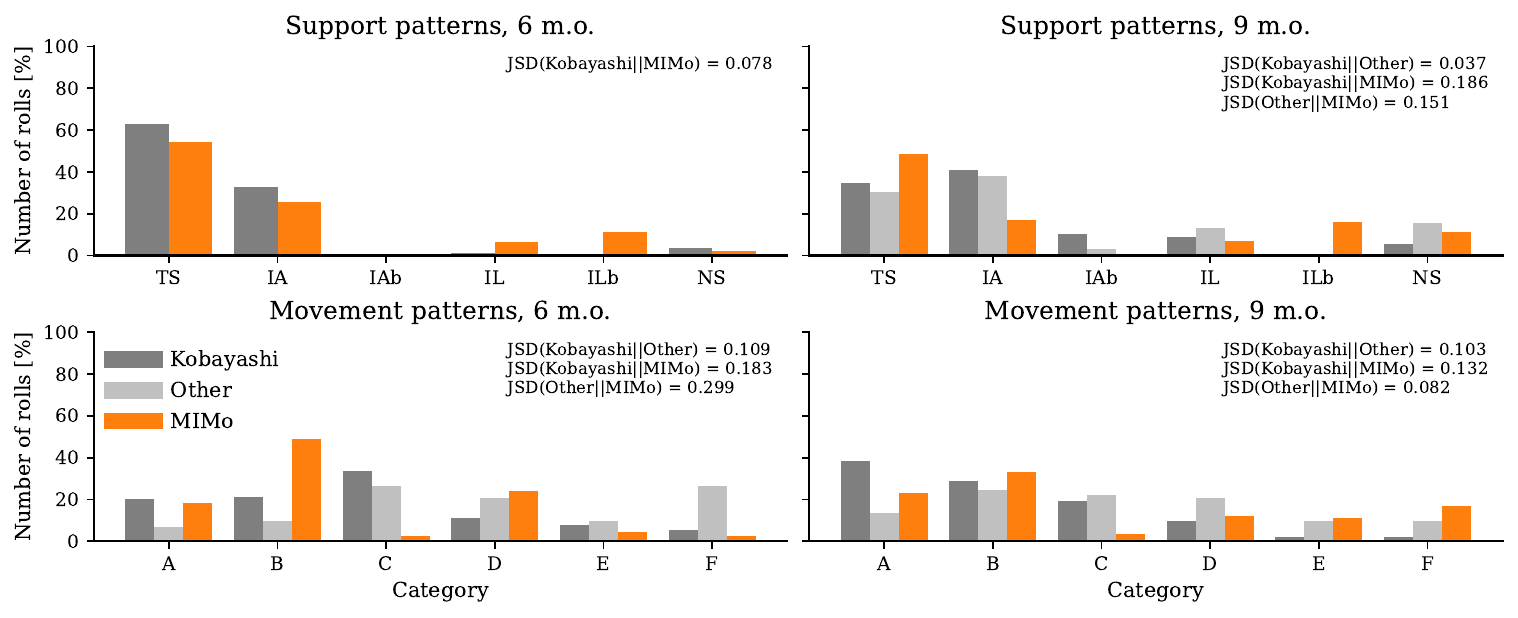}
    \caption{Distribution of support (top) and movement (bottom) patterns displayed by ``younger'' (left) and ``older'' (right) infants. Real infant data is taken from different sources. In dark gray we show the results from Kobayashi et al. \cite{kobayashi16}, who measure both coordination patterns for both age groups. In light gray are the distributions reported in \cite{kobayashi2021} for ``older'' infants (both patterns) and in \cite{siegel24:muscleactivation} for ``younger'' infants (movement patterns only). See Appendix for details about the patterns. Patterns are classified for all successful episodes within 100 attempts per model (with 18 models aged 6 months and 36 models aged 9 months). Analogous to \cite{kobayashi16}, episode data is excluded if R-squared distance of the sigmoid-fitted torso displacement is below 0.6. To quantify the similarity between MIMo and the real infants, we report the Jensen-Shannon Divergence (JSD) between pairs of distributions in each plot.} 
    \label{fig:patterns}
\end{figure*}

% \subsubsection{Disentangled cross-embodiment evaluation}
% We systematically analyze how physical strength and body size affect rolling success. We test trained models on embodiments with hybrid configurations of these variables. We initialize the tested embodiment with an asymmetric configuration, where the actuation age is derived from an embodiment of one age and the body age from another.

\subsubsection{Coordination patterns}

There exist two different approaches to categorize the coordination patterns of a roll over (see Appendix). We adapt both to MIMo and perform the categorization of test episodes automatically following the procedures decribed next. 

\paragraph{Stationary patterns}

The first and easiest approach consists of classifying a roll over according to the resulting stationary limbs \cite{kobayashi16,kobayashi2021}. To do so, Kobayashi et al. \cite{kobayashi16} calculate the time at which the torso reaches maximum displacement velocity in the direction of rolling, \(T_\mathrm{TR}\). They define a window of \(250~\mathrm{ms}\) around this time, and establish that a limb is stationary if its average velocity during that window is lower than a threshold of \(100~\mathrm{mm/s}\). We repeat this analysis with MIMo, with one critical difference. Since MIMo rolls much faster than real infants (see Fig.~\ref{fig:supine2prone}), we scale window and threshold values. The median duration of a supine-to-lateral roll for a 6-month-old MIMo is around \(500~\mathrm{ms}\), roughly 6 times faster than the values reported in \cite{kobayashi16} and \cite{siegel24:muscleactivation}. Thus, we split this difference to define a shorter window size of \(83~\mathrm{ms}\) and a higher threshold of \(300~\mathrm{mm/s}\). 

\paragraph{Movement patterns}

More refined patterns can be obtained by computing the temporal dynamics of the roll over \cite{kobayashi16,siegel24:muscleactivation,kobayashi2021}, which requires calculating the relative motion of the limbs. We adopt the methodology proposed in \cite{kobayashi16}. We fit a sigmoid function to the displacement of each moving limb to determine the time of maximum displacement velocity. This time is compared to \(T_{TR}\). Limbs that reach their peak velocity inside a window of \(83~\mathrm{ms}\) around \(T_{TR}\) are classified as ``synchronous'', whereas limbs reaching peak velocity before or after this interval are declared as ``leading'' and ``following'', respectively. As a result, the rolling motions can be classified in 6 categories, labeled A through F, as described in the Appendix.

\subsubsection{Actuation control}

Finally, drawing inspiration from the work of Siegel et al. \cite{siegel24:muscleactivation}, we analyze MIMo's actuation control during a roll over. Instead of recording muscle activity with EMG, we directly extract the actuation control commands selected for each DOF. MIMo has a total of 46 DOF, which makes the actuation analysis complex. To simplify, the actuation time series are filtered using a \(6~\mathrm{Hz}\) recursive Butterworth filter and averaged across all the control commands in the kinematic chain of each body part. For this study, we focus exclusively on the torso and four limbs. Finally, we compare the temporal dynamics of the actuation commands under different conditions.

\begin{figure*}
    \begin{subfigure}{0.24\textwidth}
        \includegraphics[width=\textwidth]{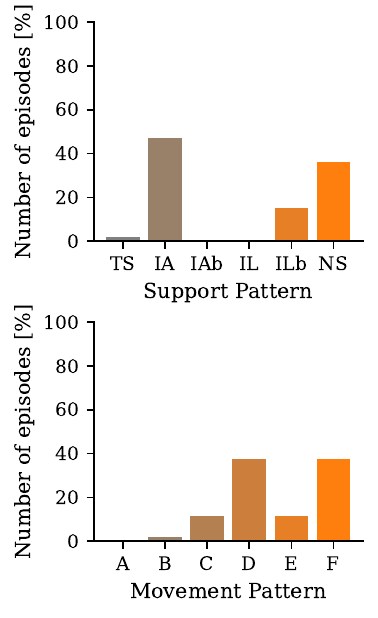}
        \caption{Distribution of patterns from a single trained model.}
        \label{fig:individual_patterns}
    \end{subfigure}
    \hfill
    \begin{subfigure}{0.74\textwidth}
        \includegraphics[width=\textwidth]{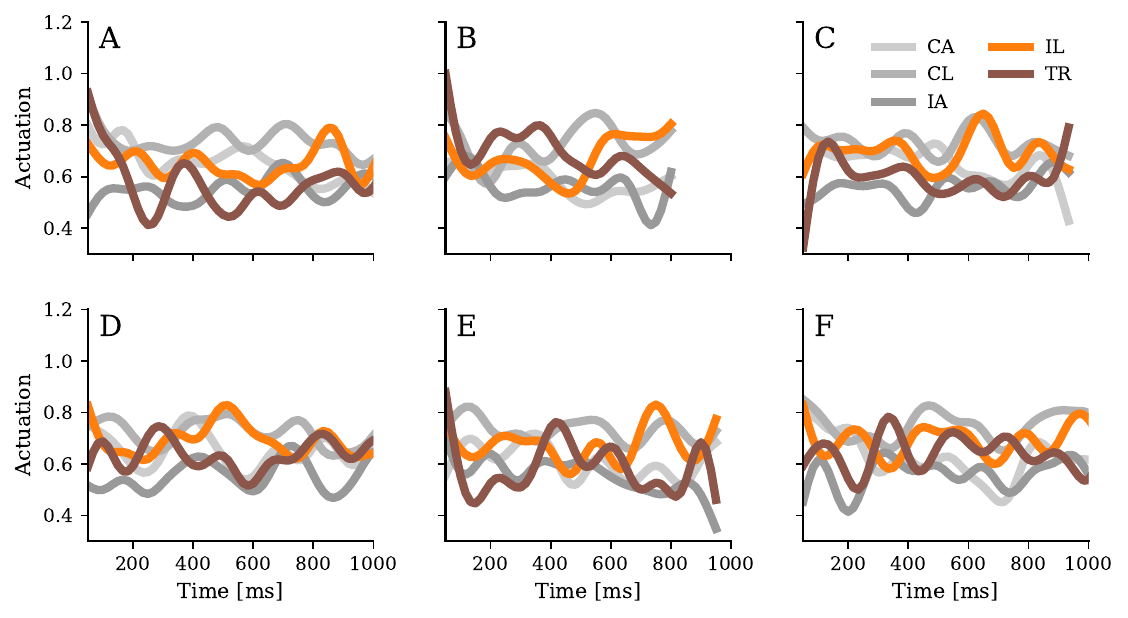}
        \caption{Examples of actuation commands over different movement patterns (A--F). Highlighted colors correspond to the ipsilateral leg (IL) and torso (TR). Other body parts are shown in gray.}
        \label{fig:actuation}
    \end{subfigure}
    \caption{Variability in coordination patterns and actuation from a single training run. (a) The model is evaluated in \(1{,}000\) test episodes with random initial conditions, and the resulting support and movement patterns are calculated for each episode. (b) Single episodes from each movement pattern are selected arbitrarily among the tests to visualize the actuation commands. These actuation time series show that a single model can show completely different actuation profiles just from the small differences in the initial conditions.}
    \label{fig:indpatternsandactuation}
\end{figure*}

\section{RESULTS}

\subsection{How embodiment constrains rolling}

Infants learn to roll somewhere between the ages of \(2\) and \(7\) months \cite{adolphmotordevelopment}, a point in development that is characterized by rapid changes in both body, brain, and experiences. The relative contributions of these three components are difficult to disentangle in behavioral experiments. Here, making use of a simulated infant like MIMo, we attempt to address this. To do so, we train the same deep reinforcement learning algorithm (brain) for the same amount of episodes (experience) but with four different embodiments, corresponding to 1, 3, 6, and 9 months of age.

The results are shown in Fig.~\ref{fig:supine2prone}. We find that increasingly older bodies allow MIMo to achieve higher success rates and roll faster. There are only two physical differences between these different models. First, MIMo's body grows to match an average infant of the same age, meaning a bigger torso, head, and limbs. Intuitively, it should be simpler to roll with a smaller (and thus lighter) body. However, body growth also entails an increase in strength and capabilities. L\'{o}pez et al \cite{mimogrows} reported that MIMo can only lift his head with a 2-month-old body and his legs with a 4-month-old body, in line with real infants. In our experiments, despite all four models being trained with the same algorithms and for the same duration, the differences in strength between the embodiments translates into large performance gaps.

To isolate the effects of body size and strength on rolling performance, we disentangle these two variables in cross-embodiment tests. We use the models trained with matching body and actuation ages and evaluate them on embodiments covering all pairs of sizes and strength. For example, this analysis allows us to test rolling performance of MIMo trained with a 1-month old embodiment on an embodiment with an actuation age of 3 months and a body age of 9 months. The results are shown in Fig.~\ref{fig:dcee}. This disentangled cross-embodiment analysis reveals that physical strength is the key to a successful roll across all body configurations. An actuation age of 1 month yields low success rates across all body sizes. Conversely, for embodiments with actuation ages of 6 or 9 months, varying the body size has little impact on rolling performance. In sum, we find that muscle development, more so than growth, supports the emergence of rolling and potentially other whole-body behaviors such as crawling or walking.

These results must only be interpreted qualitatively. As seen in Fig.~\ref{fig:supine2prone}.b, MIMo is considerably faster than real infants. That is not only because our model only has few of the many constraints and limitations of a real infant, but also because it is trained with an extrinsic reward, i.e., his only purpose is to achieve a roll over. This also explains why a fraction of the models trained with a 1-month-old embodiment achieve roll overs, something that is not observed in real 1-month-olds (although \cite{mcgraw} reported that newborns have a righting reflex that allows the to spontaneously roll to a side-lying position). Regardless, these experiments reveal how changes in the embodiment may facilitate or constrain this typical infant behavior. 

\subsection{Discovery of diverse successful movement patterns}

\begin{figure*}[!t]
    \includegraphics[width=\textwidth]{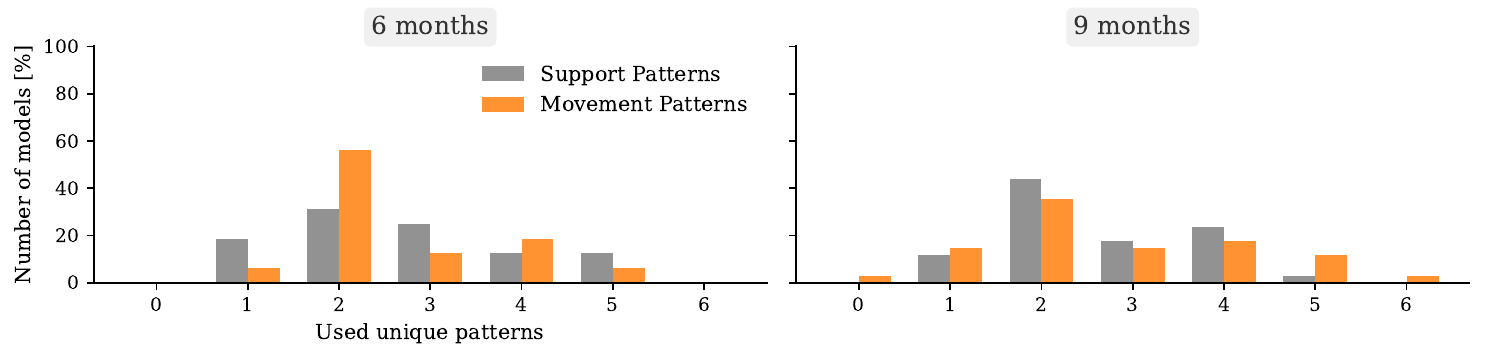}
    \caption{Distribution of number of unique coordination patterns displayed per model over 100 evaluation episodes. 18 models were trained for age 6 months and 36 for age 9 months. See Fig.~\ref{fig:patterns} and Fig.~\ref{fig:individual_patterns} for additional details.}
    \label{fig:patterns_histo}
\end{figure*}

Having found that MIMo can learn to roll, we inquire next whether the way in which he does so matches real infants. One important aspect of the rolling behavior is that there is an extensive variability of ways to achieve it. Infants have multiple rolling strategies, ranging from two stationary limbs to none \cite{kobayashi16}. We compare our model's support and movement patterns with the values reported in \cite{kobayashi16,kobayashi2021,siegel24:muscleactivation}. In those studies, infants are classified into two age groups: ``younger'' (5 and 7 months old) and ``older'' (8 to 10 months old). To match these age groups, we compare with the MIMo models trained with 6 and 9-month-old embodiments. 

The results of this comparison are shown in Fig.~\ref{fig:patterns}. We observe that MIMo, like the real infants, employs a variety of different rolling strategies. To quantify the differences in the distributions, we report the Jensen-Shannon divergence (JSD) between all pairs of distributions. A JSD of 0 means that the distributions are identical while a JSD of 1 indicates that they have no overlap. We find strong similarities between MIMo and real infants, with all JSDs being lower than 0.3. Crucially, MIMo was in no way incentivized to display a variety of coordination patterns nor was he given access to data from real infants. Hence, these results indicate that learning multiple patterns may be an efficient way to achieve roll overs from different starting positions, the only difference between evaluation episodes. 

\subsection{Adaptive behavior within a single run}

Every single model has an absolute laterality index of~\(1\) (see \cite{kobayashi16}), meaning that every trained MIMo turns either to the left or to the right, but never in both directions. This result is inherently expected when using a proximal optimization reinforcement algorithm like PPO. However, it also raises the question of whether the variability of coordination patterns described before is only a result of random model initializations rather than MIMo actually being able to learn adaptive behaviors. To address this, we arbitrarily select an individual training instance from a 9-month-old embodiment and inspect its support and movement patterns. The resulting distributions are shown in Fig.~\ref{fig:individual_patterns}. We find that this individual model utilizes different support and movement patterns depending on the initial conditions. We then extend this procedure to all the trained models, and report the number of unique patterns displayed over 100 evaluation episodes. We observe that fewer than 20\% use exactly one unique pattern, while most models display 2 patterns or more. We hypothesize that this variability within a single trained model stems from the initial conditions of the runs, such that neighboring states yield the same patterns. However, given the high-dimensional state and action spaces, we are unable to categorize different patterns according to the initial conditions.

To further investigate the origins of the variability in coordination patterns, we visualize the actuation commands used by a single model for each of the movement patterns, as shown in Fig.~\ref{fig:actuation}. We specifically compare the control of the torso and ipsilateral leg. For example, pattern movements B and D show that MIMo can adapt the relative control of his leg and torso to different scenarios. Interestingly, despite there being no \textit{a priori} incentive to learn different movements, this model executes a wide variety of coordinated actuation control sequences that explain the variability in movement patterns. Recall that the only difference from one test episode to the other is the initial condition, which results from a mild jittering of MIMo's limbs prior to the start of the episode. Nonetheless, MIMo's policy allows him to adapt to these changing scenarios.

\section{DISCUSSION}

Rolling is one of the first motor milestones of full-body coordinated control and its study can provide unique insights about early development. However, acquiring data from rolling infants is a major challenge that has slowed down progress. Alternative methods, including video recordings \cite{siegel24video} and subsequent motion retargeting analysis \cite{lopez2026simulating}, could facilitate this process. Here we have shown that MIMo, a virtual multimodal infant embodiment, can be trained to roll via reinforcement learning. The variability of coordination patterns exhibited matches qualitatively behavioral results from real infants \cite{kobayashi16,kobayashi2021,siegel24:muscleactivation}, and is explained by different combinations of actuation commands. While MIMo's learning mechanism is only a crude approximation of that of a real infant, the distribution of strategies he finds to roll shows some resemblance to the strategies used by real infants.

A closer match between our results and the reports from real infants could be achieved with increased input about resting poses. For example, it has been reported that sleep preferences can affect the latency and distributions of rolling strategies \cite{davis1998effects}. In the absence of stronger priors, the variability displayed by our model is nonetheless quite informative. MIMo is solely attempting to maximize rewards by rotating his torso. One could expect that, in spite of the random initial conditions, different training iterations would discover and converge to an identical policy, yielding a single coordination pattern. That is not the case. Thus, we speculate that the different movement sequences discovered by MIMo and by the real infants provide similarly good solutions to rolling under variable conditions.

Our main contribution has been the identification of the role of the body in the emergence of the rolling behavior. We trained the same algorithms for the same amount of time but on different embodiments, corresponding to 1, 3, 6, and 9 months of age. The results showed a clear developmental progression with higher success rates and faster movements. Crucially, by standardizing the learning conditions, we were able to identify that physical strength is a critical component of this behavior. 

There are a number of extensions that can follow this work, in particular aiming to increase the proximity of MIMo's behavior to real infants. Firstly, we will extend our experiments to more conditions, including inclined environments \cite{siegel24incline2} as well as prone-to-supine rolls. Preliminary results (not shown) anticipate that the latter may be more difficult to learn than the supine-to-prone rolls. Additionally, we will include MIMo's dynamic growth functionality during training, such that his body and actuation progressively increase with ongoing training progress. Next, to encourage exploration during training, we plan to incorporate curiosity-driven intrinsic motivations \cite{oudeyer2007intrinsic} as well as temporally auto-correlated noise \cite{lopez2026infant}, both of which are hypothesized to support learning during early infancy. Future experiments should also reflect the contributions of vision and touch. Such improvements may yield even more realistic rolling movements.

Beyond the arguments mentioned previously, a possible explanation for the difference in rolling times between real infants and MIMo may be that real infants have to deal with substantial conduction delays of sensory and motor signals from and to the periphery. Such delays are known to cause instabilities and oscillations in control systems. We speculate that young infants' relatively slow movements may help to avoid such problems. Modeling sensorimotor delays is possible within the MIMo ecosystem \cite{mimogrows} and will be incorporated in future work. Since biological development involves coupled changes in the body and the neural substrate \cite{adolph2017development}, and concretely in the myelination that determines the conduction velocities to and from the brain, MIMo can provide a suitable platform to disentangle these effects. 

 Overall, our work shows that embodied computational simulations can provide a complementary approach to the study of motor development without many of the limitations of behavioral research. The use of artificial agents opens the door for creative experimentation \textit{in silico} and allows for large scale studies of a wide variety of conditions, including atypical physical and cognitive development.

\section*{APPENDIX}

\small

\subsection*{Coordination patterns}

% The categorization of support patterns used by \cite{kobayashi16,kobayashi2021} is based on the stationary limbs. The accronyms indicate:
% \begin{description}
%     \item[\textbf{TS}] Two stationary limbs.
%     \item[\textbf{IA}] One stationary limb (IA).
%     \item[\textbf{IAb}] One stationary limb (IA) with IL moving backwards.
%     \item[\textbf{IL}] One stationary limb (IL).
%     \item[\textbf{ILb}] One stationary limb (IL) with IA moving backwards.
%     \item[\textbf{NS}] No stationary limbs.
% \end{description}

% The refined categorization of movement patterns proposed by Kobayashi et al. \cite{kobayashi16} is as follows:
% \begin{description}    
%     \item[\textbf{A}] Two stationary limbs (IA and IL) with the other two limbs (CA and CL) moving synchronously.
%     \item[\textbf{B}] Two stationary limbs (IA and IL) with the CA moving synchronously and the CL following.
%     \item[\textbf{C}] One stationary limb (IA) with the other three limbs moving synchronously.
%     \item[\textbf{D}] One stationary limb (IA) with the IL and CA moving synchronously and the CL following.
%     \item[\textbf{E}] One stationary limb (IL) with the IA and CA moving synchronously and the CL following.
%     \item[\textbf{F}] No stationary limbs with four limbs moving synchronously.
% \end{description}

The categorization of support patterns used by \cite{kobayashi16,kobayashi2021} is based on the stationary limbs. The acronyms indicate: \textbf{TS}: Two stationary limbs; \textbf{IA}: One stationary limb (IA); \textbf{IAb}: One stationary limb (IA) with IL moving backwards; \textbf{IL}: One stationary limb (IL); \textbf{ILb}: One stationary limb (IL) with IA moving backwards; \textbf{NS}: No stationary limbs.

The refined categorization of movement patterns proposed by Kobayashi et al. \cite{kobayashi16} is as follows: \textbf{A}: Two stationary limbs (IA and IL) with the other two limbs (CA and CL) moving synchronously; \textbf{B}: Two stationary limbs (IA and IL) with the CA moving synchronously and the CL following; \textbf{C}: One stationary limb (IA) with the other three limbs moving synchronously; \textbf{D}: One stationary limb (IA) with the IL and CA moving synchronously and the CL following; \textbf{E}: One stationary limb (IL) with the IA and CA moving synchronously and the CL following; \textbf{F}: No stationary limbs with four limbs moving synchronously.

\subsection*{Training hyperparameters}

\begin{figure}[!t]
      \centering
        \begin{subfigure}{0.235\textwidth}
            \centering\includegraphics[width=\textwidth]{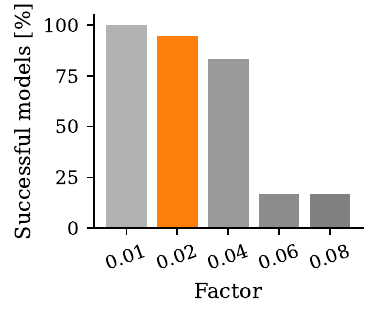}
            \caption{Penalization factor.}
            \label{fig:penalization}
        \end{subfigure}
        \hfill
        \begin{subfigure}{0.235\textwidth}
            \centering\includegraphics[width=\textwidth]{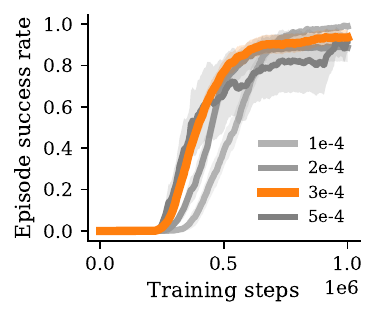}
            \caption{Learning rate.}
            \label{fig:learningrate}
        \end{subfigure}
      \caption{Training performance for different hyperparameters. Highlighted in orange are the selected values.}
      \label{fig:hyperparams}
\end{figure}

The choice of training hyperparameters used in this study was informed by an initial sweep search (see Fig.~\ref{fig:hyperparams}). We compared different penalization factors \(\alpha\) (see Section \ref{sec:reward}) and learning rates using a 9-month-old MIMo trained with supine-to-prone roll overs. The results are shown in Fig.~\ref{fig:hyperparams}. We selected \(\alpha\) as the highest value yielding more than \(90\%\) successful models (\(\alpha=0.02\) resulted in \(94\%\) successful models). We also selected a learning rate of \(3\cdot10^{-4}\) since it provided the best balance between training speed and end performance. For all other model hyperparameters we used the defaults from the Stable-Baselines3 implementation of PPO.

%                                   % the textheight of the last page by a suitable amount.
%                                   % This command does not take effect until the next page
%                                   % so it should come on the page before the last. Make
%                                   % sure that you do not shorten the textheight too much.

%%%%%%%%%%%%%%%%%%%%%%%%%%%%%%%%%%%%%%%%%%%%%%%%%%%%%%%%%%%%%%%%%%%%%%%%%%%%%%%%

%%%%%%%%%%%%%%%%%%%%%%%%%%%%%%%%%%%%%%%%%%%%%%%%%%%%%%%%%%%%%%%%%%%%%%%%%%%%%%%%

%%%%%%%%%%%%%%%%%%%%%%%%%%%%%%%%%%%%%%%%%%%%%%%%%%%%%%%%%%%%%%%%%%%%%%%%%%%%%%%%

%%%%%%%%%%%%%%%%%%%%%%%%%%%%%%%%%%%%%%%%%%%%%%%%%%%%%%%%%%%%%%%%%%%%%%%%%%%%%%%%

\bibliographystyle{IEEEtran} % use IEEEtran.bst style
\bibliography{sources}

\end{document}